\pdfoutput=1
\documentclass[11pt]{article}
\usepackage{amssymb}
\usepackage{amsfonts}
\usepackage{amsmath}
\usepackage{multirow}
\usepackage[table,xcdraw]{xcolor}

\usepackage{acl}
\usepackage{placeins}

\usepackage{times}
\usepackage{latexsym}
\usepackage{subfiles} % Best loaded last in the preamble
\usepackage{xr}
\usepackage{graphicx}
\usepackage{xurl}
\usepackage{listings}
\usepackage{paralist}
\usepackage{amsmath}
\usepackage{caption}
\usepackage{subcaption}
\usepackage[T1]{fontenc}
\usepackage[utf8]{inputenc}

\usepackage{microtype}

\title{Nonet at SemEval-2023 Task 6: Methodologies for Legal Evaluation}

\author{Shubham Kumar Nigam$^{1}$ \qquad Aniket Deroy$^{2}$ \qquad Noel Shallum$^{3}$ \\ 
\textbf{Ayush Kumar Mishra}$^{1}$ \qquad \textbf{Anup Roy}$^{1}$ \qquad \textbf{Shubham Kumar Mishra}$^{1}$ \\ \textbf{Arnab Bhattacharya}$^{1}$ \qquad \textbf{Saptarshi Ghosh}$^{2}$ \qquad \textbf{Kripabandhu Ghosh}$^{4}$\\
$^{1}$ IIT Kanpur, India \qquad
$^{2}$  IIT Kharagpur, India \\
$^{3}$ Symbiosis Law School Pune, India \qquad
$^{4}$ IISER Kolkata, India \\
\texttt{\{shubhamkumarnigam,roydanik18,noelshallum,}\\
\texttt{saptarshi.ghosh,kripa.ghosh\}@gmail.com} \\
\texttt{\{ayushkm20,puna20,skmishra20,arnabb\}@iitk.ac.in}\\
}%author

\begin{document}

%% Shubham
\newcommand\SN[1]{\textbf{\textcolor{red}{(#1)$_{Shubham}$}}}

%% Aniket
\newcommand\Aniket[1]{\textcolor{orange}{(#1)$_{Aniket}$}}

%% Ayush
\newcommand\Ayush[1]{\textcolor{purple}{(#1)Ayush}}

\maketitle
 
\begin{abstract}
This paper describes our submission to the SemEval-2023 for Task 6 on LegalEval: Understanding Legal Texts. Our submission concentrated on three subtasks: Legal Named Entity Recognition (L-NER) for Task-B, Legal Judgment Prediction (LJP) for Task-C1, and Court Judgment Prediction with Explanation (CJPE) for Task-C2. We conducted various experiments on these subtasks and presented the results in detail, including data statistics and methodology. It is worth noting that legal tasks, such as those tackled in this research, have been gaining importance due to the increasing need to automate legal analysis and support. Our team obtained competitive rankings of 15$^{th}$, 11$^{th}$, and 1$^{st}$ in Task-B, Task-C1, and Task-C2, respectively, as reported on the leaderboard.
%\SN{Decide title of the paper}
\end{abstract}

\section{Introduction}
The SemEval Task-6~\cite{legaleval-2023} aims to automate several tasks to streamline the Indian legal process, which is often slow and delayed due to the country's large population and a shortage of judicial resources. Additionally, people in India are not always fully aware of the country's laws. To make the legal process more accessible to the general public, the SemEval task addresses crucial problems that are specific to the Indian judicial context. One task is the "Legal Named Entity Recognition (L-NER)" system, which identifies named entities in the legal text. Legal judgments contain intriguing entities like the names of the petitioner, respondent, judge, lawyer, date, organization, GPE, statute, provision, precedent, case number, witness, and other persons, which are typically not recognized by conventional entity recognition systems. Therefore, developing systems that are tailored to the legal domain is crucial. In task B, participants were tasked with identifying the legal entities present in legal judgments. A court judgment is divided into two parts: the preamble, which includes the names of the parties, the court, lawyers, and other details, and the decision (judgment), which follows the preamble. The organizers separately provided the preamble and judgment text datasets. 

Apart from the Legal NER system, the SemEval Task-6 also includes another important task $C$, that is crucial for automating the Indian legal process. The subtask, referred to as task C1, is the Legal Judgment Prediction (LJP) task. This task aims to determine whether the legal judgment favors the appellant or the defendant, which is modeled as a binary classification problem. Given a large number of pending legal cases in India, automating the process of predicting legal judgments can significantly reduce the burden on the judicial system. Moreover, the task of finding an explanation for the legal decision is equally important. For this, we locate a span from the legal judgment that is highly correlated with the reasoning behind the legal decision. Subtask C2 focuses on finding a suitable explanation for the binary classification task in subtask C1, thereby providing an additional layer of transparency to the legal process.

To accomplish these tasks, we experiment with several models and techniques. For the Legal Named Entity Recognition system, we try out a Spacy-based model and a fine-tuned BERT model to detect the legal named entities. For the Legal Judgment Prediction task, we pass the last 512 tokens of the legal judgment through transformer models, as well as try out hierarchical transformer models on the entire dataset to train the models for the task of judgment prediction. Lastly, for the Court Judgment Prediction with Explanation task, we check various span lengths taken from the end of the document. The datasets provided to the participants as a component of the SemEval task were in the English language.

Our main contributions can be summarized as:\\
(1) Our contributions in this paper include participating in three subtasks: Legal Named Entity Recognition (L-NER), Legal Judgment Prediction (LJP), and Court Judgment Prediction with Explanation (CJP). \\
(2) In the L-NER task, we explored three models: BERT-CRF, modified spaCy pipeline, and transformer embeddings. \\
(3) For the LJP task, we experimented with various hierarchical transformer models and utilized pre-trained transformer models on the last 512 tokens of judgments. \\
(4) For the subtask of Explanation for Prediction, we proposed an intuitive approach of keyword-based matching technique for court judgment decision identification and extracted the court judgment explanation using the span lengths from the ending portions of legal judgments. \\
 We released the codes and datasets for all subtasks via GitHub\footnote{\url{https://github.com/ShubhamKumarNigam/LegalEval23_Nonet}}.
% \vspace{-0.3cm}
\section{Background}
%We participated in task B-Legal Named Entity recognition, C1-Legal Judgment Prediction, and C2-Explanation for Prediction. 
We have participated in task-B named Legal Named Entity Recognition (L-NER), task-C1 called Legal Judgment Prediction (LJP), and task-C2, called Court Judgment Prediction with Explanation (CJPE).

% The sub-task, L-NER is formulated as labeling each token with a class for each named entity and a class named "0" for tokens that do not contain any entities. The input for this task is a sentence, and the output is the annotated text with named entities. For the sub-task of legal judgment prediction, the input is the legal judgment, and the outcome is the binary label-accept or reject. So the task of legal judgment prediction is a binary classification task. For the sub-task of explanation for prediction, the input is a court judgement and the output is a relevant text span from the document which contributes to the decision.
% We now discuss in details all the related works:-\\
\textbf{Legal Named Entity Recognition:}
The task of Legal Named Entity Recognition (Legal NER)~\cite{skylaki2020named} involves identifying named entities in structured legal text. These entities include petitioner, respondent, court, statute, provision, precedent, and more. However, conventional Named Entity Recognizers, like SPACY, do not recognize these entity types, highlighting the need for domain-specific legal NER systems. Additionally, the Indian legal system has its unique processes and terminologies, making it necessary to develop a specific legal NER for Indian court decision texts. Unfortunately, there are no publicly available annotated datasets for the task of L-NER on Indian court cases.

To address this gap, a recent study~\cite{kalamkar2022named} has developed a Legal NER model specifically for Indian legal data. This work highlights the need for an annotated dataset to train the Legal NER model and proposes a methodology for creating such a dataset. By training their model on the created dataset, they achieved state-of-the-art results on the task of Legal NER on Indian legal data.

% Legal NER~\cite{skylaki2020named} is the task of detecting and identifying named entities in structured text. Legal documents have entities such as petitioner, respondent, court, statute, provision, precedent, etc. The conventional Named Entity Recognizer, like SPACY, does not recognize these entity types. As a result, a domain-specific legal named entity recognizer is required. Moreover, there are no publicly available annotated datasets for Indian courts. Due to the uniqueness of Indian legal processes and terminologies, it is necessary to create a separate legal NER for Indian court decision texts. There is a work~\cite{kalamkar2022named} which develops a Legal NER model on Indian legal data. There are no publicly available annotated datasets for the task of L-NER on Indian Court cases.

\textbf{Legal Judgment Prediction:}
Legal judgment prediction is an important task in the field of legal informatics. A recent work by Malik et al.~\cite{malik-etal-2021-ildc} focuses on predicting the outcome of Indian Supreme Court cases using state-of-the-art transformer models. The task is modeled as a binary classification problem and applied to the entire legal case except for the final judgment. The authors use the manually annotated explanations for every legal case judgment in the Indian Legal Document Corpus (ILDC) dataset as gold standards for the court judgment explanation task.

Another work that addresses the task of judgment prediction is the Chinese AI and Law Challenge dataset~\cite{xiao2018cail2018}. This dataset is the first large-scale Chinese legal dataset with 2.6 million cases, which makes it an excellent resource for judgment prediction on Chinese data. The human annotation in this work is rich, and the authors compared simple text classification approaches, such as FastText, TFIDF+SVM, CNN, etc., on the facts of the legal judgment.

The Hindi Legal Document Corpus~\cite{kapoor2022hldc} is a corpus of 900K legal documents annotated in Hindi, designed for the task of bail prediction. The work experiments with a range of state-of-the-art models and proposes a multi-task learning framework where the main task is bail prediction, and the auxiliary task is summarization.

Finally, there is a work on Legal judgment prediction~\cite{chalkidis-etal-2019-neural}, which provides a dataset of 11k legal judgments. The work compares different models, including BiGRU-attn, HAN, LWAN, BERT, and Hier-BERT, on the facts of the legal case.

\textbf{Explainability Methods:}
The method of Integrated Gradients is an explainability method~\cite{sundararajan2017axiomatic} for deep learning models where we have to load a trained deep learning model into this method, and then the method of Integrated Gradients computes the gradient of the output given by the model to its input features. The model gives attribution scores to every sentence in a document. The method helps in understanding and extracting the features which contribute to a  model decision.
Also, there are standard explainability techniques like LIME and SHAP for machine learning models. SHAP~\cite{lundberg2017unified} is a predominant explainability technique that focuses on the individual impact of every feature on the model's final prediction.
LIME~\cite{ribeiro2016should} tries to understand how the perturbations in the model input affect the final output prediction.
Explainable AI~\cite{10.1145/3404835.3462799} is a recent and vibrant research area gathering a lot of focus. The work presents a recent explainable system called SIMFIC 2.0 which is an enhanced version of a recent explainable system. The idea here is to define the notion of similarity in fiction books. The system uses handcrafted interpretable features for fiction books and then provides a global explanation for fitting a linear regression and local explanation based on similarity features.
Explainable Information retrieval~\cite{anand2022explainable} is an emerging research area that tries to improve upon the trustworthiness of the information retrieval methods.
% \vspace{-0.3cm}
\section{System Overview}
In this section, we provide an overview of the methodologies used for the tasks we participated in, which include Legal Named Entity Recognition (L-NER), Legal Judgment Prediction (LJP), and Court Judgment Prediction with Explanation.
\subsection{Legal Named Entity Recognition (L-NER)}
\subsubsection{Using SpaCy Model}
For L-NER, we propose an innovative approach that leverages the strengths of pre-trained transformer models and domain-specific embeddings to build custom NER models in the legal domain. Specifically, our model architecture involves fine-tuning a RoBERTa ~\cite{liu2019roberta} transformer with Spacy\footnote{https://spacy.io/} and incorporating external embeddings from Law2Vec\footnote{https://archive.org/download/Law2Vec} (200 dimensions) to enhance the corpus's contextual understanding of legal terms.

\begin{figure}[h]
  \includegraphics[width=\linewidth]{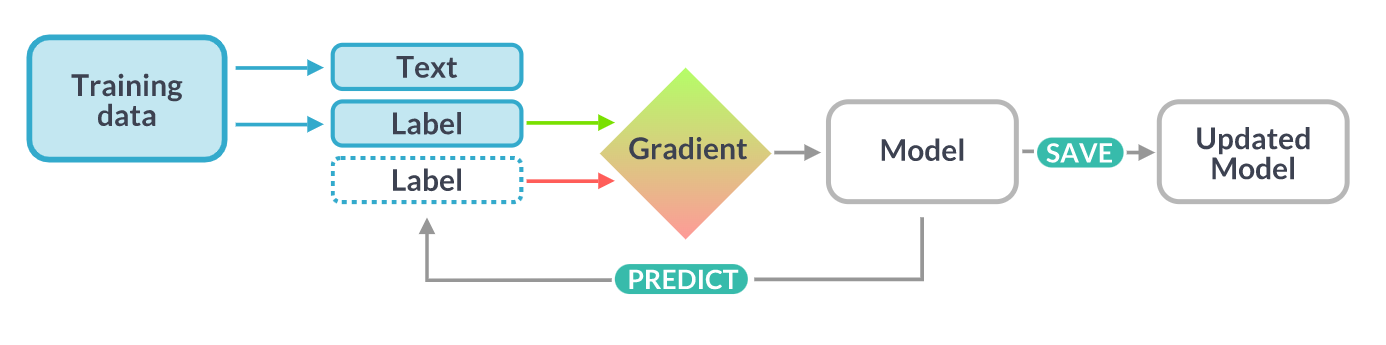}
  \caption{Spacy Custom NER Training Procedure }
  \label{fig:Spacy Training Procedure}
\end{figure}

Our pipeline involves data preparation, pipeline configuration, model fine-tuning, and evaluation. To begin with, we use the dataset provided by the organizers and create Doc objects that contain the text. We also incorporate external embeddings from Law2Vec to enhance the corpus's contextual understanding of legal terms. The custom Spacy's model training procedure is shown in Figure~\ref{fig:Spacy Training Procedure}\footnote{https://spacy.io/usage/training/}, which involves an iterative process of comparing the model's predictions against reference annotations to estimate the gradient of the loss and using it to calculate the gradient of the weights through back-propagation. The gradients indicate how the weight values should be changed so that the model’s predictions become more similar to the reference labels over time.

\FloatBarrier

The Spacy pipeline configuration includes the RoBERTa transformer model, the Law2Vec embeddings, and additional NER components using the transformer, ner, and vectors components provided by Spacy. We fine-tune the BERT transformer model with the Adam.V1 optimization algorithm, a learning rate of $2e^{-5}$, and 100 epochs, and evaluate the performance of our model on a separate test set.
SpaCy Model description in detail is here\footnote{https://tinyurl.com/SpaCy-Model-description}. SpaCy is powerful but somewhat opaque and hard to modify.
Overall, our work demonstrates the effectiveness of our model architecture for fine-tuning BERT transformers with Spacy 3 for Legal NER and highlights the potential of incorporating external embeddings from Law2Vec for building high-performance NER models in the legal domain.

\subsubsection{Fine-Tuned BERT Model}
\begin{figure}[h]
  \includegraphics[width=\linewidth]{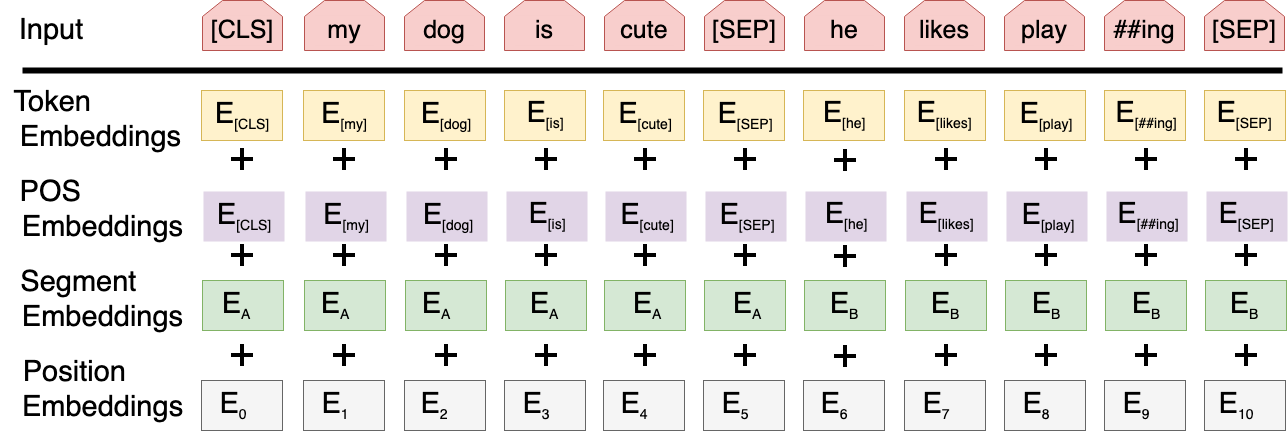}
  \caption{Modified BERT Embedding Layer}
  \label{fig:Modified BERT Embedding Layer}
\end{figure}

We perform the task of Named Entity Recognition for legal documents using a pre-trained legal BERT base uncased model~\cite{chalkidis-etal-2020-legal}. The model architecture comprises a BERT model with a token-level classifier, followed by a Linear-Chain CRF. \\
For an input sequence of n tokens, BERT outputs an encoded token sequence with hidden dimension H. The classification model takes each token's encoded representation to the tag space, i.e., $\mathbb{R}^H$ -> $\mathbb{R}^K$, where $K$ is the number of tags in the dataset. And the output score maps to $\mathbb{R}^{n*K}$ of the classification model which is then fed to the CRF layer, which is used to predict in BIO format.\\
When we tested on BERT-CRF, our model could have performed better in this way because the sentence length was too much and the number of entities significantly less. Even in some of the training sentences, there were no entities in the sentence. We tried to make it more robust by adding part of the speech tag in the embedding layer of the BERT model. The observation behind providing further Embedding is that most of the named entities are nouns, so we want our span to be more exact.\\
In the classical BERT model ~\cite{devlin-etal-2019-bert}, we have input embeddings as the sum of the token embeddings, the segment embedding, and the position embedding. After getting input embedding, it passes to Bert Layer Norm, then the dropout layer. After that, we return to the Embedding.\\
Actually, we cannot directly append a POS tag with the token because the pre-trained tokenizer only knows the token, not the token + POS tags. A better way is to create an additional input to the model(besides input\_ids and token\_type\_ids) called pos\_tags\_ids, for which we add an additional embedding layer(nn. Embedding) ~\ref{fig:Modified BERT Embedding Layer}. In that way, we sum the Token embeddings, token types, and POS tags. The max number of a pos tag is the total unique number of POS tags, also called the "vocabulary size" of the embedding layer. The hidden size is the size of the embedding vector that we want to learn for each pos tag(which is 768 by default for BERT-base). 
In order to improve the performance of the BERT-CRF model for named entity recognition (NER), we have generated part-of-speech (POS) tags as an additional feature for the provided dataset. These tags are used in conjunction with the tokens during training to make NER tag predictions. The POS tags are generated for each token in the sentences using a pre-trained spaCy~\cite{spacy2} model, which has demonstrated an accuracy of 97\%\footnote{https://spacy.io/usage/facts-figures} for this specific task.  

An additional complexity of BERT-like models is that they rely on subword tokens, rather than words. This means that a word like "playing" might be tokenized into ["play", "\#\#ing"]. This means that we will also have to provide POS tags at the token level. Similar to how each token is turned into an integer (input\_ids), we will also have to turn each POS tag into a corresponding integer (pos\_tag\_ids) in order to provide it to the model. So we would actually need to keep a dictionary that maps each POS tag to a corresponding integer.

\subsubsection{Fusion of Model}
As discussed above, two models were trained to perform legal name entity recognition task. These models are SPACY and BERT-CRF. The output of both models is an array of tuples consisting of the entity's starting index, ending index, and label. To determine the final prediction, if the tuples in the output of both models intersect (start, end), then the intervals are merged, keeping the label the same for the final prediction. For instance, if the prediction from the SPACY model is (43,53, ORG) and that of the BERT-CRF model is (45,60, ORG), the final output will be (43,60, ORG). Otherwise, the union of outputs is taken from both models. This approach improved the prediction accuracy by 2\% compared to using only the SPACY model for this task.

\begin{figure}[h]
  \includegraphics[width=\linewidth]{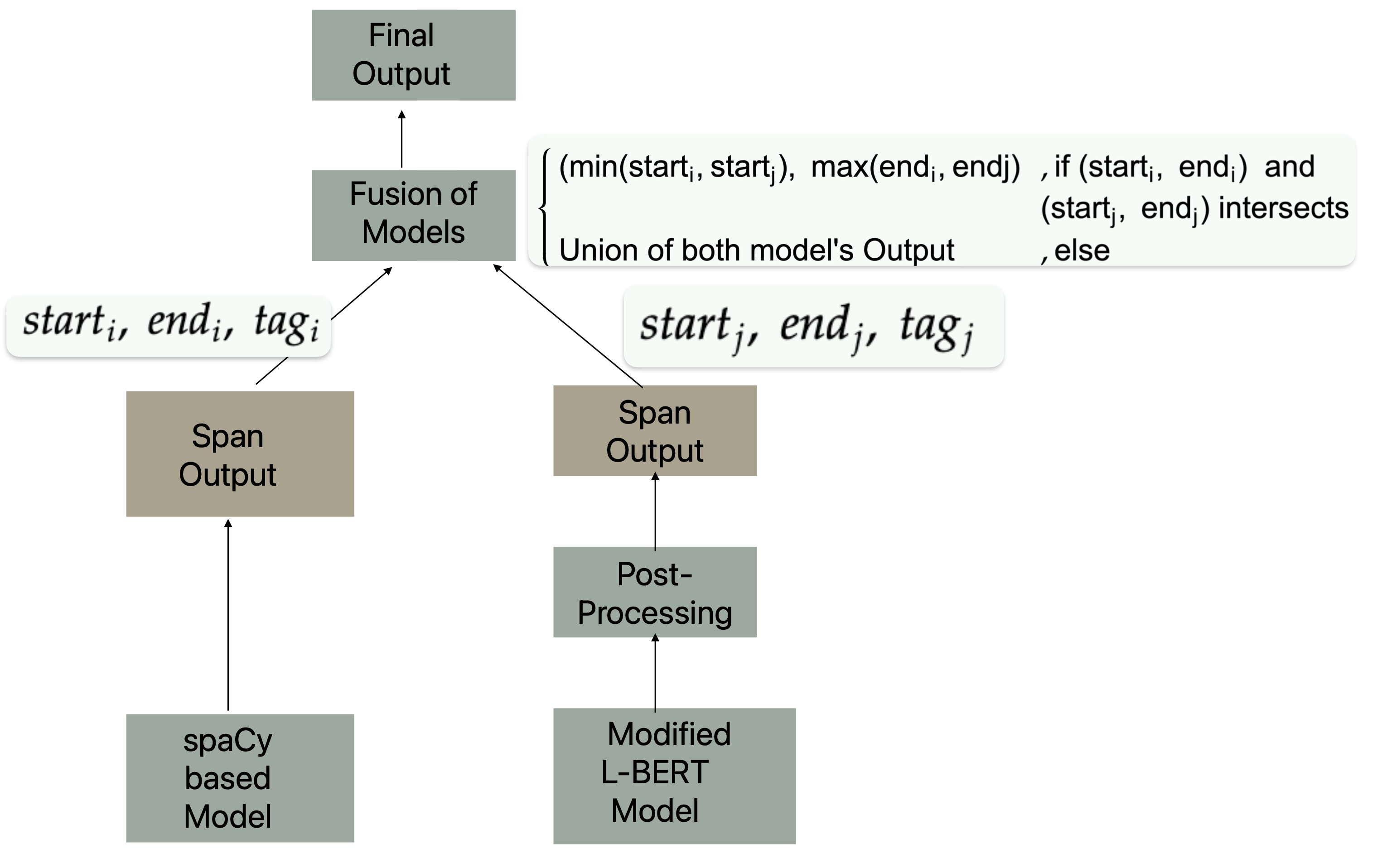}
  \caption{Fusion of Models}
  \label{fig:models_fusion}
\end{figure}
% \FloatBarrier

\subsection{Court judgment Prediction with Explanation (CJPE)}

Task C was divided into two sub-tasks:
\begin{enumerate}
    \item Legal Judgment Prediction (LJP):  given a legal judgment document, the task involves automatically predicting the outcome (binary: accepted or denied).
    \item Explanation for Prediction: explanations are in the form of relevant text spans in the document that contributes to the decision.
\end{enumerate}

\begin{figure*}[!t]
    \centering
    \frame{\includegraphics[width=\linewidth]{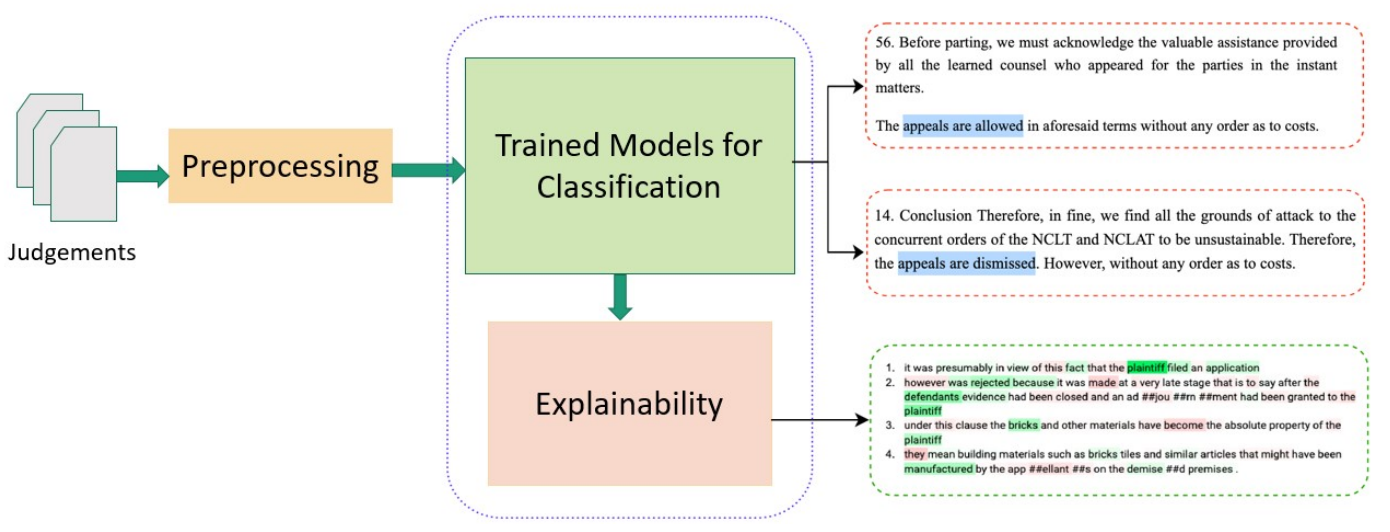}}
    \caption{Court Judgment Prediction and Explanation (Task C) schematic diagram}
    \label{fig:task-c-schematic-dia}
\end{figure*}

For better understanding, we make the schematic diagram \ref{fig:task-c-schematic-dia} for court judgment prediction and explanation. In that diagram, first, we preprocess the documents where we remove the meta information of case documents like judge name(s), court name, petitioner name(s), defendant name(s), hyperlinks, etc. (if it is present). Then pass, it to transformer-based models to classify the outcome. We pass either the part of case documents that contain the relevant information or chunked the case file with an overlap of 100 token window size. Then the second subtask is to give or highlight the spans of text in the document, which helps or contributes to the decision.

%%%%%%%%%%%%%%%%%%%%%%%%%%%%%%%%%%%%%%%%%%%%%%%%%%%%%%%%%%%%%%%%%%%%%%%%%%%%%%%%%%%%%%%%

%%%%%%%%%%%%%%%%%%%%%%%%%%%%%%%%%%%%%%%%%%%%%%%%%%%%%%%%%%%%%%%%%%%%%%%%%%%%%%%%%%%%%%%%

\subsubsection{Legal Judgment Prediction (LJP)}
We used the pre-trained transformers which are trained on general corpus like XLNet \cite{NEURIPS2019_dc6a7e65} and Roberta \cite{DBLP:journals/corr/abs-1907-11692} as well as trained on legal corpus LegalBERT~\cite{chalkidis-etal-2020-legal}, InlegalBERT, IncaseLawBERT~\cite{paul-2022-pretraining} and we also train the BERT large on Indian judgment cases. Since the transformers have restrictions that they can not accommodate more than 512 tokens, so we gave only the last 510 tokens (two special tokens are reserved for CLS and SEP) as \cite{malik-etal-2021-ildc} mentioned in the paper that in general most relevant information is present at the end of the documents. Another way to accommodate the long document is that we took inspiration from \cite{chalkidis-etal-2020-legal} and we also tried Hierarchical Transformer model architecture. We divided each document into chunks using a moving window approach where each chunk was of length 512 tokens, and there was an overlap of 100 tokens. We obtained the [CLS] representation of these chunks, which were then used as input to sequential models (BiGRU + attention). But we are not getting better accuracy on hierarchical models compared to the transformer-based models. The possible reason could be provided data is not sufficient for passing the embedding information to the sequential models. 

\subsubsection{Explanation for Prediction}
We perform the task of Court judgment prediction by following pattern matching. We check for keywords that represent whether a court judgment is in favor of the appellant or against the appellant. We check for keywords namely dispose of, disposed of, accept, allow, allowed, accepted, and upheld in the entire court judgment which are keywords that help in detecting decisions in favor of the appellant. We check for keywords namely dismiss, dismissed, discard, discarded, reject, and rejected in the court judgment which are keywords that help in detecting decisions against the appelant.%So keyword based matching used to check decisions in favour or against the apellant.

There are research works~\cite{polsley-etal-2016-casesummarizer} on extractive legal document summarization where the legal domain experts say that towards the end of a legal judgment, there are sentences that tend to summarize the entire judgment.

Current research works on abstractive summarization of legal judegements~\cite{salaun2022conditional} also suggests that the legal judgment is reversed and then fed into abstractive encoder-decoder architectures like BART which can take inputs up to a fixed input length of 1024 tokens so that the important information present at the end of a judgment is not missed.

\citet{malik-etal-2021-ildc} states that the most important information corresponding to the court judgment is present towards the end portions of the legal document. The most important syntactic and semantic information occurs toward the end of a case. The largest occlusion and attention scores are assigned to the chunks present at the end of the document.
%We know that the last portion of a court judgment contains information regarding the explanation of the judgment of a legal case. 
For explaining the judgment of a court, we need to explain the reason behind the court judgment, which is given by the judges and generally present towards the end of a court judgment. The beginning portions of a legal document do not contain information that gives you a suitable explanation for explaining the court judgment decision.
%he explanations are in the form of relevant text spans in the document that contribute to the decision. 
The problem description says that the explanations should be in form of relevant text spans in the document that contributes to the decision. So we take up a simple and intuitive approach to address the problem of finding a suitable explanation for the court judgment. So we choose continuous text spans of different span lengths from the end of the legal judgment. So we capture the last 550 words, last 520 words, last 512 words, last 500 words, last 450 words, last 400 words, last 350 words, last 300 words, and last 250 words of every court judgment document to act as an appropriate explanation of a court judgment.

For the purpose of breaking down every court judgment text into words, we used the split() function available in Python.

Though there are various sophisticated explainability approaches like the Method of Integrated Gradients which can rank the sentences present in the court judgment on the basis of attribution scores. These methods tend to work well if the model trained on the legal judgment prediction task has got high f1-scores. If the legal judgment prediction model trained on legal judgments is not giving good f1-scores on the test set, then the sophisticated explainability methods(like the Method of Integrated Gradients) might not work well because the legal judgment prediction model itself is not very accurate. The pre-trained XLNET-large model trained on the last 512 tokens on the train set has given a low f1-score of 0.5287 on the test set(for subtask C1). So we can understand that the pre-trained XLNET large model trained for subtask-C1 is not very accurate. So we decided to follow a simple yet intuitive approach for the task of Court judgment prediction with explanation rather than using a sophisticated explainability approach(like the Method of Integrated Gradients).
Any other standard explainability approach will also not work well because the best-performing model XLNET large( on subtask C1) has a low f1 score of 0.5287 on the test set and hence using this model any standard explainability approach will not be able to give a good explanation for the prediction task. 
% \vspace{-0.3cm}
%\begin{document}
\section{Experimental Setup}
%\textcolor{red}{Needs to be done for all subtasks.}
\subsection{Legal-Named Entity Recognition (L-NER)}

%\section{Dataset}
\subsubsection{Dataset Description of L-NER}
\begin{figure}
  \includegraphics[width=\linewidth]{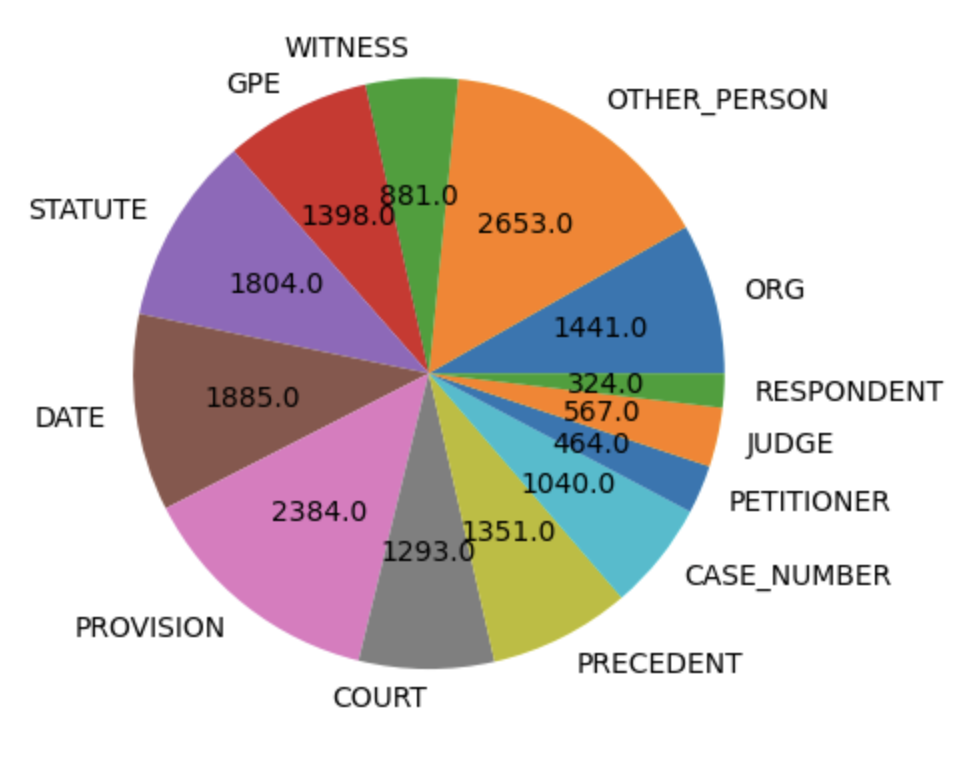}
  \caption{Judgement Tag Distribution}
  \label{fig:judgment_piechart}
\end{figure}

\begin{figure}
  \includegraphics[width=\linewidth]{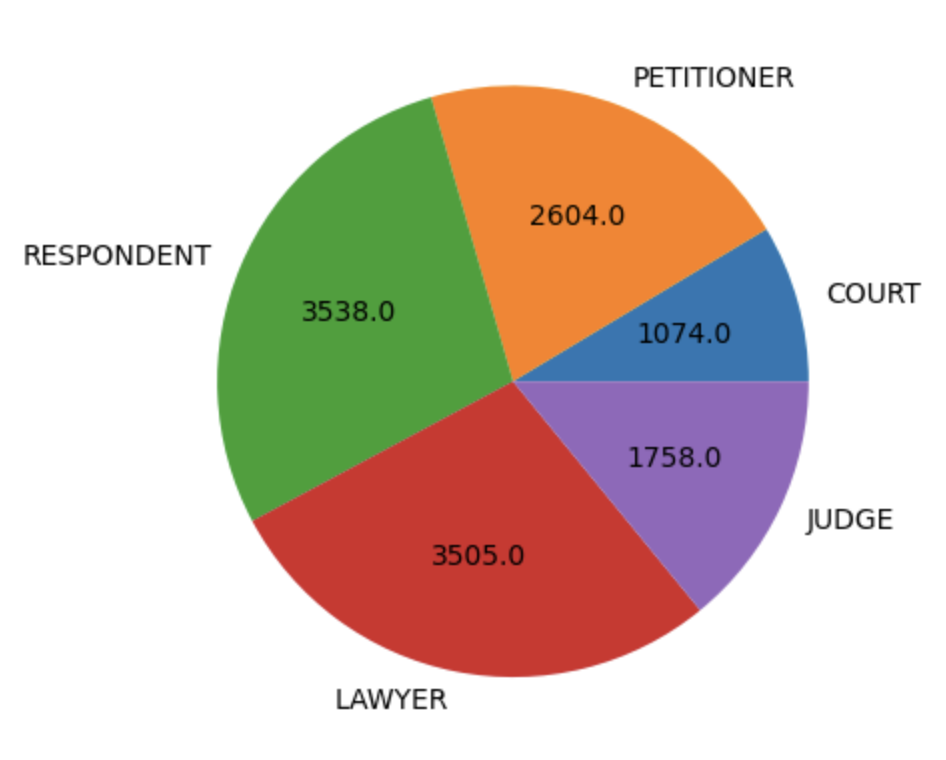}
  \caption{Preamble Tag Distribution}
  \label{fig:preamble_piechart}
\end{figure}

The Dataset provided comprised two sections. The first one was the Preamble, which contained the name of the parties, court, lawyers, etc., and the second one was the judgment, which contained the name of the court, case number, precedent, provision, witness, etc. There were 13 tags in the Judgment section and 5 in the Preamble section. The distribution of tags for the two sections is shown in Figure ~\ref{fig:judgment_piechart} and Figure ~\ref{fig:preamble_piechart} respectively, and the distribution of sentences for the Train, Dev, and Test Data split is shown in the Table ~\ref{tab:data_distribution}.

\begin{table}[h]
% \tiny
\centering
\resizebox{0.9\columnwidth}{!}{
\tiny
% \scriptsize
% \large
\begin{tabular}{|c|c|c|c|}
\hline
          & \textbf{Train}     & \textbf{Dev} & \textbf{Test} \\ \hline
\textbf{PREAMBLE}  & 1560 & 125   & 441 \\ %\hline
\textbf{JUDGMENT} & 9435 &  949 & 4060 \\ \hline
\end{tabular}}
\caption{\label{tab:data_distribution}
Train, Dev and Test Data Split for L-NER
}
\end{table}

\subsubsection{Dataset Preprocessing for L-NER}
Different preprocessing layers are involved in L-NER, which consists of preprocessing before running the model, which requires cleaning the dataset in which extra spaces are removed (between words and start and end of sentences), similarly removing extra symbols, i.e., repetition of non-alphanumeric characters. We observed that there were multiple newline characters, tabs, and extra spaces in the dataset because of the structure of the legal document of India. After the preprocessing, there occurred differences in spans of entities in the original text and preprocessed text, so we used regex to find new start and end indexes for entities in preprocessed text for the SPACY model, and for the BERT model, we used BIO format for training over preprocessed text.
At the time of prediction, for the BERT model, we converted the BIO format output back to the JSON format, which had the entities with their labels and starting and ending indexes in the sentence.\\

% For the final prediction on the test dataset, we have combined the results produced by both the SPACY and BERT-CRF model, i.e., if the span of entities predicted intersects, then we take the entity with a larger span. For example, if the predicted entity by the SPACY model is "HDFC Bank" and by the BERT-CRF model is "HDFC Bank, Noida" ,  we take "HDFC Bank, Noida" as the final prediction. For the rest of the predictions, we take the union of the result produced by both models. By this technique, we observed better accuracy on the test dataset.
\subsubsection{Hyperparameters of L-NER}
Table \ref{tb:hyperparameter for l-ner} shows the hyperparameters for both the models which gave best result for this task.
\begin{table}[h]
\centering
% \tiny
\footnotesize
\label{tab:hyperparameters}
\begin{tabular}{|p{2.9cm}|p{1.3cm}|p{2cm}|}
\hline
\textbf{Hyperparameter} & \textbf{SPACY} & \textbf{L\_Bert-Crf} \\
\hline
Learning Rate & $5e^{-4}$ & $3e^{-5}$ \\
% \hline
Batch Size & 128 & 32 \\
% \hline
Number of Epochs & 50 & 50 \\
% \hline
% Number of Hidden Layers & 2 & 3 \\
% \hline
% Hidden Layer Sizes & [64, 32] & [128, 64, 32] \\
% \hline
Activation Function & ReLU & ReLU \\
% \hline
Dropout Rate & 0.1 & 0.25 \\
% \hline
Optimizer & Adam & AdamW \\
% \hline
Loss Function & Cross-Entropy & Cross-Entropy \\
\hline
\end{tabular}
\caption{Hyperparameters for the Models for L-NER}
\label{tb:hyperparameter for l-ner}
\end{table}
% We performed hyper-parameter tuning of each model and selected the best hyper-parameter giving the best F-score on validation set. Model training parameters are divided into two groups with different learning rate: $1e^{-3}$ and $4^{e-3}$ for fine tuning L-BERT model and $3e^{-5}$ for hidden layers in L-BERT CRF with 100 number of epochs, and  We have used different set of batch size {4, 8, 16, 32}. Similarly, we vary the learning range {$1e^{-3}$, $3e^{-3}$, $4e^{-3}$, $5e^{-3}$, $1e^{-4}$, $3e^{-4}$, $5e^{-4}$, $1e^{-5}$, $5e^{-5}$} and a Adam optimizer with weight decay of $1e^{-3}$ and epsilon value as $1e^{-8}$.
% % \input{files/ner_dataset.tex}

%%%%%%%%%%%%%%%%%%%%%%%%%%%%%%%%%%%%%%%%%%%%%%%%%%%%%%%%%%%%%%%%%%%%%%%%%%%%%%%%%%%
\subsection{Legal Judgment Prediction (LJP)}

\subsubsection{Dataset for LJP}
The provided dataset for task-C1, Legal judgment Prediction (LJP) consisted of two sets of ILDC$_{single}$, and ILDC$_{multi}$. The ILDC$_{single}$ and ILDC$_{multi}$ contain 4982 and 5082 train documents, correspondingly. For both sets, organizers provided 994 and 1500 common validation and test documents correspondingly. Data split is shown in Table \ref{tab:task-b-dataset}.

%%%%%%%%%%%%%%%%%%%%%%%%%%%%%%%%%%%%%%

\begin{table}[]
\centering
\resizebox{0.9\columnwidth}{!}{%
\tiny
\begin{tabular}{|l|l|l|l|}
\hline
\textbf{Corpus} & \textbf{Train} & \textbf{Dev} & \textbf{Test} \\ \hline
ILDC\_single    & 4982           & 994  & 1500        \\ %\hline
ILDC\_multi     & 5082           & 994  & 1500        \\ \hline
\end{tabular}%
}
\caption{LJP statistics}
\label{tab:task-b-dataset}
\end{table}

%%%%%%%%%%%%%%%%%%%%%%%%%%%%%%%%%%%%%%%%%%%%%%%%%%%%%%%%%%%%%%%%%%%%%%%%%%%%%%%%%%%

\subsubsection{Hyperparameters of LJP}
If the document is longer than 10000 tokens, then we truncate the document. In transformers, we kept the batch size 16, epochs = 5, learning rate 2e-6, and the remaining parameters set to default. 

For hierarchical transformers we make a chunk size of 500 tokens and an overlapping of 100 tokens. 

\subsection{Dataset for Explanation for Prediction} 
%This is an unsupervised task called C2:Court judgment prediction with explanation. 
The dataset for task-C2, court judgment prediction with explanation consisted of 50 legal court judgments.
%The task is to find out the court judgment and a corresponding explanation to why the judgment is accepted or denied.
The average no of words in the dataset is 2315.12 words. The average no of sentences in the dataset is 207.98. We tried to detect the rhetorical roles in the dataset of 50 court judgments. 
%The evaluation measure for Court judgment prediction task is macro-F1 score and the evaluation measure for Court judgment explanation task is Rouge-2 f1 score. 

Now we are trying to analyze the dataset in terms of rhetorical roles to get a better understanding of the nature of the dataset.

\begin{table}[h]
\centering

\small{
\begin{tabular}{|c|c|}
\hline
\textbf{Rhetorical role} & \textbf{Fraction of rhetorical role}  \\ \hline
Facts & 0.3858  \\% \hline
Argument & 0.0425  \\% \hline
Ruling by present court & 0.2284  \\ %\hline
Ruling by lower court & 0.0124  \\ %\hline
Precedent & 0.1270  \\ %\hline
Statute & 0.1975  \\ %\hline
Ratio of the decision & 0.0061  \\ \hline
\end{tabular}}

\caption{\label{tab:rhetorical_role_courtjudgmentexplanation}
Fraction of Rhetorical roles in every document in the dataset of 50 court judgments
}
%}
\end{table}
Table~\ref{tab:rhetorical_role_courtjudgmentexplanation} shows the fraction of Rhetorical roles~\cite{ghosh2019identification} in every document in the dataset of 50 court judgments. The work~\cite{ghosh2019identification} of rhetorical role detection uses a hierarchical BiLSTM-CRF model to address the problem of rhetorical role detection. The hierarchical BiLSTM layers extract necessary features from the sentence and the CRF layer helps to design the sequential presence of sentences belonging to different rhetorical roles.
%The sentences belonging to Ratio of the decision, Ruling by present court and Ruling by Lower court are generally considered very important for providing an explanation of the case judgment. %The sentences belonging to these rhetorical roles mostly tend to occur in the ending portions of the document.
We have run the rhetorical role extractor just to understand and analyze the nature of the dataset. But we have not used sentences belonging to specific rhetorical roles to select a text span from the court judgment as an explanation for the court judgment. Basically, the rhetorical role extractor has just been used to give us a better understanding of the nature of the dataset.
%\end{document}
% \vspace{-0.3cm}
%\documentclass[11pt]{article}
%\usepackage{xr}
%\addbibresource{anthology.bib}

%\begin{document}
\section{Results and Analysis}
\subsection{Legal Named Entity Recognition (L-NER)}
% We ranked 15th in the L-NER as shown in Table \ref{tab:subtask-b-leaderboard}, i.e., Task-B of the competition. Our F1-Score is 0.5532 on the test dataset. During training, we used 80\% of the provided training data for training the model and the remaining 20\% as validation data. The organizer's validation data was used as a test dataset to evaluate the precision, recall, and F1 score of our model. We selected our best-performing model for submission in this task. Our model's performance is reflected in the classification report over the validation data as a test dataset. L-NER F1-SCORE on preamble Table \ref{tab:SCORE_FOR_PREAMBLE} validation data, which we used as test data 85.50. L-NER F1-SCORE on judgment Table \ref{tab:SCORE_FOR_judgment} validation data that we used as test data 83.74.

In Task-B of the competition, we attained a ranking of 15th place, as demonstrated in Table \ref{tab:subtask-b-leaderboard}. Our model achieved an F1-Score of 0.5532 on the test dataset. Our model was trained using 80\% of the available training data, with the remaining 20\% utilized as validation data. To evaluate the precision, recall, and F1 score of our model, the organizer's validation data was utilized as the test dataset. The best-performing model was chosen for submission. The classification report over the validation data serves as an indication of our model's performance on the test dataset. The weighted average F1-Score on the validation data, used as the test dataset, was 85.50 and 83.74 for the preamble and judgment, respectively, as illustrated in Tables \ref{tab:SCORE_FOR_PREAMBLE} and \ref{tab:SCORE_FOR_judgment}.  

There is a significant difference between the F1-Score achieved on the test dataset compared to what was displayed in the classification report for the validation dataset. Unfortunately, as the labeled test dataset has not been released, we are unable to provide a comprehensive analysis of why the model did not perform as well as it did on the validated dataset used as the test dataset.

\begin{table}[h]
\resizebox{\columnwidth}{!}{%
\begin{tabular}{|c|c|c|c|}
\hline
\textbf{Rank}& \textbf{User} & \textbf{Team Name} & \textbf{F1} 
\\ \hline

1                              & Pinal-Patel                    & ResearchTeam\_HCN                   & 0.912                        \\ %\hline
2                              & bluesky                        & -                                   & 0.9099                       \\ %\hline
3                              & DeepAI                         & -                                   & 0.9099                       \\ %\hline
\textbf{15}                    & \textbf{ShubhamKumarNigam}     & \textbf{Nonet}                      & \textbf{0.5532}              \\ \hline
\end{tabular}%
}
\caption{Scores and leader-board ranks for subtask B}
\label{tab:subtask-b-leaderboard}
\end{table}

%-----------------------------------------------------------------------------------

\begin{table}[h]
        \centering
        \resizebox{\columnwidth}{!}{
        \tiny
        \begin{tabular}{|c|c|c|c|}
        \hline
              \textbf{Entity type}& \textbf{Precision} & \textbf{Recall} & \textbf{F1-Score}\\
             \hline
               COURT & 83.15 & 97.37 & 89.70\\
               PETITIONER & 65.50 & 90.34 & 75.94\\
               LAWYER & 88.74 & 90.13 & 89.43\\
               RESPONDENT & 78.95 & 79.71 & 79.33\\
               JUDGE & 93.53 & 89.04 & 91.23\\
               \hline
               \multicolumn{3}{|c|}{Weighted Average} & 85.50\\
            \hline
        \end{tabular}}
        \caption{\label{tab:SCORE_FOR_PREAMBLE}L-NER per Entity Score for PREAMBLE on Validation dataset}   
    \end{table}
% -------------------------------------------------------------------------------------
    \begin{table}[h]
        \centering
        \resizebox{\columnwidth}{!}{
        \tiny
        \begin{tabular}{|c|c|c|c|}
        \hline
             
              \textbf{Entity type}& \textbf{Precision} & \textbf{Recall} & \textbf{F1-Score}\\
             \hline
               STATUTE & 90.61 & 90.61 & 90.61\\
               PRECEDENT & 68.97 & 76.92 & 72.73\\
               JUDGE & 72.73 & 100.00 & 84.21\\
               GPE & 77.91 & 74.16 & 75.64\\
               OTHER PERSON & 86.72 & 88.68 & 87.69\\
               DATE & 82.85 & 98.51 & 90.00\\
               PROVISION & 88.39 & 89.59 & 88.99\\
               CASE NUMBERS & 75.59 & 84.96 & 80.00\\
               COURT & 91.72 & 89.60 & 90.64\\
               ORG & 67.69 & 59.86 & 63.54\\
               PETITIONER & 50.00 & 77.78 & 60.87\\
               WITNESS & 90.00 & 93.10 & 91.53\\
               RESPONDENT & 100.00 & 80.00 & 88.89\\
               \hline
               \multicolumn{3}{|c|}{Weighted Average} & 83.74\\
            \hline
        \end{tabular}}
        \caption{\label{tab:SCORE_FOR_judgment}L-NER per Entity Score for JUDGEMENT on Validation dataset}
        \label{tab:len}
    \end{table}
% ------------------------------------------------------------------------------------

\subsection{Legal Judgment Prediction (LJP)}
For task C-1 metric used for evaluation is the standard Macro F1 score. We ranked 11$^{th}$ with 4 entries in the LJP task and got a 0.5287 Macro F1 score using a pre-trained XLNet large model. The result of our best model is in table \ref{tab:subtask-c1-leaderboard} along with the top 3 teams that participated in this task. In the training phase, we tried different pre-trained transformers which are trained on the general corpus and trained on the legal corpus. We also tried hierarchical transformers but we get the best result by using the XLNet model.

\begin{table}[h]
\resizebox{\columnwidth}{!}{%
\begin{tabular}{|c|c|c|c|}
\hline
\textbf{Rank}& \textbf{User} & \textbf{Team Name} & \textbf{F1} 
                                   \\ \hline
1                              & bluesky                        & -                                   & 0.7485                       \\ %\hline
2                              & irit\_iris                     & irit\_iris                          & 0.7228                       \\ %\hline
3                              & uottawa.NLP23                  & uottawa.nlp23                       & 0.6782                       \\ %\hline
\textbf{11}                    & \textbf{ShubhamKumarNigam}     & \textbf{Nonet}                      & \textbf{0.5287}              \\ \hline
\end{tabular}%
}
\caption{Scores and leader-board ranks for subtask C-1}
\label{tab:subtask-c1-leaderboard}
\end{table}

%%%%%%%%%%%%%%%%%%%%%%%%%%%%%%%%%%%%%%%%%%%%%%%%%%%%%%%%%%%%%%%%%%%%%%%%%%%%%%%%%%%%%%

In Table \ref{LJP on training dataset}, we present the results of the Legal Judgment Prediction (LJP) task. We experimented with various pre-trained transformers, including those trained on general and legal corpora. Among these models, the best performance was achieved with XLNet\_large. We also explored hierarchical transformers; however, we did not observe improved accuracy compared to transformer-based models. This could be due to the lack of sufficient data for effectively passing embedding information to sequential models.

%%%%%%%%%%%%%%%%%%%%%%%%%%%%%%%%%%%%%%%%%%%%%%%%%%%%%%%%%%%%%%%%%%%%%%%%%%%%%%%%%%%%%%

\begin{table}[h]

        \centering
        \resizebox{\columnwidth}{!}{
        \begin{tabular}{|c|c|c|c|c|}
        \hline
              \textbf{Models}& \textbf{Precision} & \textbf{Recall} & \textbf{F1-Score}\\
             \hline
InLegalBERTLarge                                              & 0.6825                                                                 & 0.6821                                                              & 0.6823                                                          \\% \hline
LegalBert                                                     & 0.6299                                                                 & 0.6288                                                              & 0.6294                                                          \\ %\hline
\textbf{XLNet\_large}                                                  & \textbf{0.7703}                                                                 & \textbf{0.7596}                                                              & \textbf{0.7649 }                                                         \\ %\hline
Roberta\_large                                                & 0.7580                                                                 & 0.7354                                                              & 0.7465                                                          \\ %\hline
InLegalBERT                                                   & 0.7468                                                                 & 0.7364                                                              & 0.7416                                                          \\ %\hline
InCaseLaw                                                     & 0.7162                                                                 & 0.7072                                                              & 0.7117    
\\ \hline
\end{tabular}}
\caption{Judgment prediction results on Validation dataset}
\label{LJP on training dataset}
\end{table}

%%%%%%%%%%%%%%%%%%%%%%%%%%%%%%%%%%%%%%%%%%%%%%%%%%%%%%%%%%%%%%%%%%%%%%%%%%%%%%%%%%%%

\subsection{Explanation for Prediction}

For task C-2 metric used for evaluation is the standard ROUGE-2 score.

% ------------------------------------------------------------------------------------

\begin{table}[h]
\resizebox{\columnwidth}{!}{%
\begin{tabular}{|c|c|c|c|c|}
\hline

\textbf{Rank}& \textbf{User} & \textbf{Team Name} & \textbf{F1}  & \textbf{ROUGE-2}
                                   \\ \hline
\textbf{1}                              & \textbf{ShubhamKumarNigam}              & \textbf{Nonet}                               & \textbf{0.5417}                       & \textbf{0.0473}                            \\ %\hline
2                              & bluesky                        & -                                   & 0.4797                       & 0.047                             \\ %\hline
3                              & nicolay-r                      & nclu\_team                          & 0.4789                       & 0.0465                            \\ \hline
\end{tabular}%
}
\caption{Scores and leader-board ranks for subtask C-2.}
\label{tab:subtask-c2-leaderboard}
\end{table}

% ------------------------------------------------------------------------------------

We ranked in the first position in the Court judgment prediction with explanation subtask. We show the leaderboard of task C2: Court judgment prediction with an explanation for the set of 50 court judgments in table~\ref{tab:subtask-c2-leaderboard}. The macro-f1 score for the court judgment prediction task is 0.5417.

%We know that the last portion of a court judgment contains information regarding the explanation of the judgment of a legal case. For explaining the judgment of a court judgment we need to explain the reason behind the court judgment which is given by the judges and generally present towards the end of a court judgment.

Table~\ref{tab:results_courtjudgmentexplanation} shows the results of Court judgment prediction subtask and explanation subtask for 50 court judgments. The highest rouge-2 score for the explanation subtask is obtained for the last 300 words for all court judgments. 
%The lowest rouge-2 score is obtained for last 550 words for all court judgments. 

\begin{table}[!htbp]
\centering

\resizebox{1.0\columnwidth}{!}{
\begin{tabular}{|c|c|c|}
\hline
%\textbf{Rhetorical role} & \textbf{Fraction of rhetorical role}  \\ \hline
\textbf{Explanation span length} & \textbf{Rouge-2 Score} & \textbf{Percentage of Text covered}\\ \hline
Last 250 words & 0.0458 & 16.05 \\% \hline
\textbf{Last 300 words} & \textbf{0.0473} & 19.26\\% \hline
Last 350 words & 0.0468 & 22.47\\ %\hline
Last 400 words & 0.0462  & 25.69\\ \hline %\hline
%Last 450 words & 0.0456  \\ %\hline
%Last 500 words & 0.0448  \\ %\hline
%Last 512 words & 0.0446  \\ %\hline
%Last 520 words & 0.0447 \\ %\hline
%Last 550 words & 0.0442 \\ \hline
\end{tabular}}

\caption{\label{tab:results_courtjudgmentexplanation}
Results of task C2: Explanation for Prediction for the set of 50 court judgments. The best result is indicated in bold. Also, the percentage of text covered in the entire court judgment(measured in terms of no. of words) for different span lengths for all 50 court judgments is mentioned.
}
%}
\end{table}

% ------------------------------------------------------------------------------------

%The figure~\ref{fig:span_length} shows the span length in the x-axis which implies explanation of the legal case whereas the y-axis represents the rouge-2 scores.

% ------------------------------------------------------------------------------------

%\begin{table}[!htbp]
%\centering

%\resizebox{\columnwidth}{!}{
%\begin{tabular}{|c|c|}
%\hline
%\textbf{Rhetorical role} & \textbf{Fraction of rhetorical role}  \\ \hline
%\textbf{Explanation span length} & \textbf{Percentage of text covered} \\ \hline
%Last 250 words & 16.05 \\ %\hline
%\textbf{Last 300 words} & \textbf{19.26} \\ %\hline
%Last 350 words & 22.47 \\ %\hline
%Last 400 words & 25.69 \\ \hline %\hline
%Last 450 words & 28.90 \\ %\hline
%Last 500 words & 32.11 \\ %\hline
%Last 512 words & 32.88 \\ %\hline
%Last 520 words & 33.39 \\ %\hline
%Last 550 words & 35.32 \\ \hline
%\end{tabular}}

%\caption{\label{tab:results_percentage}
%This table shows percentage of text covered in the entire court judgment(measured in terms of no of words) for different span lengths for all 50 court judgments.
%}
%}
%\end{table}

% ------------------------------------------------------------------------------------

Also, Table~\ref{tab:results_courtjudgmentexplanation} shows the percentage of text covered in the entire court judgment(measured in terms of no of words) for different span lengths for all 50 court judgments. The best results are obtained for the last 300 words, and the percentage of text covered is 19.26\%.

% \textbf{Error Analysis:} To conduct an error analysis, it is necessary to evaluate the strengths and weaknesses of our proposed methods and techniques for subtasks C1, C2, and B. Unfortunately, we cannot perform an error analysis on the test dataset as the organizers have not yet released the gold standard data corresponding to the test set for every subtask. Error analysis involves comparing the outputs generated by our methods/techniques with the gold standard output to identify the flaws in our system. As a result, we are unable to conduct a comprehensive error analysis.

% \textbf{Error Analysis}- Error analysis involves looking at the strengths and weaknesses of our proposed methods and techniques for the subtask C1, subtask C2 and subtask B. But the gold standard data corresponding to the test set for every subtask has still not been released by the organizers. So it is not possible for us to perform error analysis on the test dataset because error analysis involves comparing the outputs provided by our methods/techniques with the gold standard output to find out the flaws in our system.
%\end{document}

%\begin{figure}[ht]
%\centering
%\includegraphics[scale=0.27]{files/hits(17).pdf}
%\caption{Rouge-2 for explanation vs Span Length. X-axis represents the span length. Y-axis represents Rouge-2}
%\label{fig:span_length}
%\end{figure}

% ------------------------------------------------------------------------------------

% \vspace{-0.3cm}
\section{Conclusion}
In conclusion, we participated in three subtasks of the SemEval-2023 for Task 6 on LegalEval: Legal Named Entity Recognition (L-NER), Legal Judgment Prediction (LJP), and Court Judgment Prediction with the explanation. For L-NER, we utilized a combination of pre-trained transformer models, domain-specific embeddings, and a modified spaCy pipeline to build a high-performing NER model. In LJP, we experimented with various hierarchical transformer-based models and utilized the last 512 tokens of judgments through pre-trained transformer models. For the explanation task, we found that taking the last 300 words from the end of the legal judgment performed best in terms of the rouge-2 f1 score.

Furthermore, we compared the results of different models for each subtask to analyze their strengths and weaknesses. However, since the gold standard data corresponding to the test set for every subtask was not released by the organizers, we were unable to perform an error analysis on the test dataset. We will do this once the gold standard labels are available publicly.

Looking ahead, we plan to explore joint optimization frameworks that incorporate both rhetorical role information and sentence position in the document to create more accurate span detection models. Overall, our participation in the competition has allowed us to showcase the effectiveness of our proposed models and techniques for tasks in the legal domain.

% We tried out various hierarchical transformer-based models for the task of legal judgment prediction along with passing the last 512 tokens of the judgments through pretrained transformer models. For the explanation task, we tried out various span lengths taken from the end of the legal document and we found out that the last 300 words taken from the end of the legal judgment performed best in terms of rouge-2 f1 score. Also for the L-NER task, we tried out the BERT-CRF model which is a hierarchical model as well a spacy-based model. We have tried to analyze the results of different models by comparing the results of different models with one another. 
% Currently, we have used only the location of words present in a court judgment to detect appropriate spans from the court judgment. In the future, we can jointly learn to use both the rhetorical role information and the position of the sentences in the document to create a joint optimization framework for better span detection.
% \vspace{-0.3cm}
\newpage
% \section*{Acknowledgments}
% We express our sincere gratitude to the anonymous reviewers for their diligent and insightful feedback. We have made every effort to incorporate their valuable suggestions and improve the quality of this article. The meticulous reviewing process has undoubtedly contributed to the refinement of our work.
% We would like to thank the anonymous reviewers for sincere and hard-worked reviews. We have tried our level best to improve the quality of the article. Our article got improved because of the careful reviewing process.
\bibliography{anthology,custom}

% \appendix

% \section{Appendix}
% \label{sec:appendix}

% \input{files/todo.tex}

\end{document}